\documentclass{article}
\usepackage{arxiv}
\usepackage[utf8]{inputenc}
\usepackage[T1]{fontenc}
\usepackage{hyperref}
\usepackage{url}
\usepackage{booktabs}
\usepackage{amsfonts}
\usepackage{amsmath}
\usepackage{nicefrac}
\usepackage{microtype}
\usepackage{cleveref}
\usepackage{graphicx}
\usepackage{natbib}
\usepackage{doi}
\usepackage{multirow}
\usepackage{array}
\usepackage{subcaption}
\usepackage{listings}
\usepackage{xcolor}
\usepackage{tabularx}
\usepackage{makecell}
\usepackage{float}
\usepackage[section]{placeins}
\makeatletter
\renewcommand\section{\@startsection{section}{1}{\z@}%
  {-0.9ex \@plus -.15ex \@minus -.15ex}%
  {0.15ex \@plus .05ex}%
  {\normalfont\large\bfseries}}
\renewcommand\subsection{\@startsection{subsection}{2}{\z@}%
  {-0.7ex \@plus -.15ex \@minus -.15ex}%
  {0.05ex \@plus .05ex}%
  {\normalfont\normalsize\bfseries}}
\renewcommand\subsubsection{\@startsection{subsubsection}{3}{\z@}%
  {-0.6ex \@plus -.1ex \@minus -.1ex}%
  {0.05ex \@plus .05ex}%
  {\normalfont\normalsize\bfseries}}
\makeatother

\renewcommand{\headeright}{}
\renewcommand{\undertitle}{}
\title{Sentiment Analysis of AI Adoption in Indonesian Higher Education
       Using Machine Learning and Transformer-Based Models}
\author{%
  Happy Syahrul Ramadhan\\
  Faculty of Science\\
  Sumatra Institute of Technology\\
  Lampung, Indonesia\\
  happy.122450013@student.itera.ac.id
  \And
  Ahmad Sahidin Akbar\\
  Faculty of Science\\
  Sumatra Institute of Technology\\
  Lampung, Indonesia\\
  ahmad.122450044@student.itera.ac.id
  \AND
  Karin Yehezkiel Sinaga\\
  Faculty of Science\\
  Sumatra Institute of Technology\\
  Lampung, Indonesia\\
  karin.123410029@student.itera.ac.id
  \And
  \textbf{Luluk Muthoharoh} \\
  Faculty of Science\\
  Institut Teknologi Sumatera \\
  Lampung Selatan, Indonesia \\
  luluk.muthoharoh@sd.itera.ac.id \\
  \AND
  \textbf{Ardika Satria} \\
  Faculty of Science\\
  Institut Teknologi Sumatera \\
  Lampung Selatan, Indonesia \\
  ardika.satria@sd.itera.ac.id \\  
  \And
  Martin C.T. Manullang\\
  Faculty of Industrial Technology\\
  Sumatra Institute of Technology\\
  Lampung, Indonesia\\
  martin.manullang@if.itera.ac.id\\
}

\renewcommand{\shorttitle}{Sentiment Analysis of AI Adoption in Indonesian Higher Education}
\hypersetup{
  pdftitle  = {Sentiment Analysis of AI Adoption in Indonesian Higher Education Using Machine Learning and Transformer-Based Models},
  pdfsubject= {Natural Language Processing, Sentiment Analysis},
  pdfauthor = {Happy Syahrul Ramadhan, Ahmad Sahidin Akbar, Karin Yehezkiel Sinaga, Martin C.T. Manullang},
  pdfkeywords={Sentiment Analysis, Natural Language Processing, DistilBERT,
               Support Vector Machine, Indonesian Higher Education,
               Artificial Intelligence},
}

\begin{document}
\maketitle

\begin{abstract}
Artificial Intelligence (AI) adoption in Indonesian higher education has prompted both support and concern, especially regarding learning behavior and academic integrity. This study analyzes Indonesian student opinions using TF-IDF-based machine learning and Transformer-based deep learning. The dataset contains 2,295 labeled samples, combining 1,154 student opinions with additional lexical sentiment data. LightGBM, Random Forest, and Support Vector Machine (SVM) are evaluated in the machine learning setting, while DistilBERT is fine-tuned for binary classification. SVM achieves the best machine learning performance with 82.14\% test accuracy and F1-score, while DistilBERT performs best overall with 84.78\% accuracy and 84.75\% F1-score. These results show the advantage of contextual Transformer representations, although SVM remains attractive for efficient sentiment classification.
\end{abstract}

\keywords{Sentiment Analysis \and Natural Language Processing \and DistilBERT \and
          Support Vector Machine \and Indonesian Higher Education \and
          Artificial Intelligence}

\section{Introduction}
\label{sec:intro}

Artificial Intelligence (AI) is increasingly being adopted in higher education, ranging from adaptive learning systems \citep{liu2012sentiment} and virtual assistants to generative AI tools such as ChatGPT \citep{vaswani2017attention}. These technologies offer significant opportunities to improve learning efficiency, enhance access to information, and support greater academic flexibility.

Despite these benefits, the growing use of AI in educational settings also raises important concerns. Issues such as misuse, overdependence on AI-generated outputs, and the potential decline of students' critical thinking skills have become central topics of discussion \citep{medhat2014sentiment}. Understanding how students perceive the use of AI is therefore essential for evaluating both its opportunities and its risks in higher education.

One useful way to examine these perceptions is through sentiment analysis, which enables the systematic identification and classification of opinions expressed in text. Sentiment analysis can help institutions better understand students' attitudes and use this information to support more informed academic policies and decision-making. This is particularly relevant for Indonesian-language data, where the availability of suitable NLP resources has continued to improve \citep{wilie2020indonlu}.

This study compares two different approaches to sentiment classification: a classical machine learning model and a deep learning model. Support Vector Machine (SVM) with TF-IDF is employed as the baseline model \citep{medhat2014sentiment}, while DistilBERT \citep{sanh2019distilbert} is used to represent the Transformer-based approach, which is known for its stronger contextual language representation \citep{devlin2019bert}. In general, SVM is computationally efficient and performs well on traditional text classification tasks, whereas Transformer-based models often achieve higher accuracy by capturing richer semantic context.

By leveraging Indonesian NLP resources such as IndoNLU \citep{wilie2020indonlu}, this study aims to compare the performance of SVM--TF-IDF and DistilBERT in analyzing student opinions regarding the use of AI in higher education. The findings are expected to contribute to a better understanding of suitable sentiment analysis methods for Indonesian educational data and to provide insights into how students respond to the growing integration of AI in academic environments.

\section{Related Work}
\label{sec:related}

This section reviews classical sentiment analysis methods, Transformer-based models, and the
Indonesian NLP context relevant to this study.

\subsection{Classical Approaches for Sentiment Analysis}

Sentiment analysis identifies the polarity of opinions in text \citep{sutoyo2022prdect}.
Classical approaches typically use vector representations and classifiers such as Naive
Bayes, Logistic Regression, Random Forest, or SVM, and remain useful as simple, stable
baselines.

\subsection{TF-IDF Representation}

TF-IDF is a standard text representation that highlights terms that are important within a
document but less common across the corpus \citep{salton1988term}. It remains effective with
models such as SVM in efficient and reproducible experiments.

\subsection{Support Vector Machine}

SVM is a maximum-margin classifier widely used in text classification
\citep{cortes1995support}. It handles high-dimensional TF-IDF features well and remains a
strong baseline because of its robustness, efficiency, and consistent performance.

\subsection{Transformer Models and Indonesian Context}

Transformers use self-attention to capture contextual relationships more effectively than
earlier architectures \citep{vaswani2017attention}. These models are widely used when
contextual representation and predictive performance are important \citep{wolf2020transformers}.

\subsection{Student Perceptions of AI and ChatGPT in Higher Education}

The emergence of ChatGPT has intensified discussions on the use of generative AI in
education. Previous studies show mixed responses, including concerns about academic
integrity and output reliability \citep{teerakapibal2025chatgpt}. These findings highlight
the need for similar studies in more specific local contexts. This study addresses that
need by comparing SVM--TF-IDF and Transformer-based approaches on opinion data from
Indonesian university students.

\subsection{BERT and DistilBERT}

BERT extends the Transformer architecture through bidirectional pre-training
\citep{devlin2019bert}, while DistilBERT preserves most of BERT's performance at lower
computational cost \citep{sanh2019distilbert}.

\subsection{Indonesian NLP Context}

Indonesian NLP has advanced through resources such as IndoNLU \citep{wilie2020indonlu} and
NusaCrowd \citep{cahyawijaya2023nusacrowd}. Even so, studies on Indonesian students'
perceptions of AI in higher education remain limited. ChatGPT has intensified debate on
generative AI in education, especially around academic integrity and output reliability
\citep{teerakapibal2025chatgpt}. This makes local evidence important, including studies on
Indonesian university students.

\section{Methodology}
This section outlines data collection, preprocessing, data splitting, modeling, and evaluation.

\subsection{Research Workflow}
\label{sec:research_workflow}

\begin{figure}[H]
  \centering
  \includegraphics[width=0.28\textwidth]{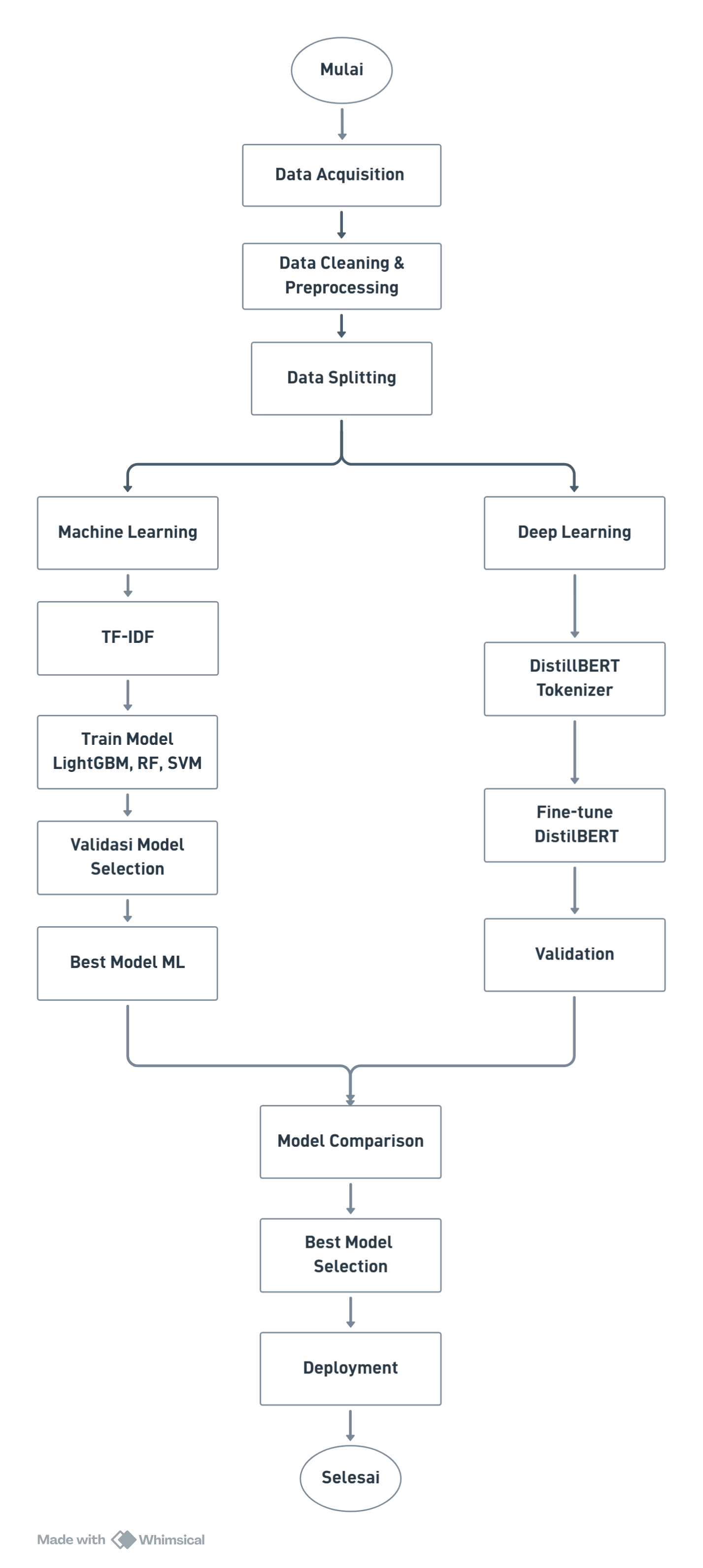}
  \caption{Research workflow.}
  \label{fig:research_workflow}
\end{figure}

Figure~\ref{fig:research_workflow} summarizes the workflow. The study collects Indonesian
student opinions on AI in higher education, preprocesses the text, and splits the dataset
into training, validation, and test sets. Two branches are then applied: TF-IDF with
LightGBM, Random Forest, and SVM, and a fine-tuned DistilBERT model. The models are
compared using accuracy, precision, recall, and F1-score.

\subsection{Dataset}

The dataset combines 1{,}154 Indonesian student opinions on AI in higher education with
lexicon-based positive and negative sentiment data. After merging both sources, the final
dataset contains 2{,}295 samples for binary sentiment classification.

Details of the data sources used in this study are presented in Table~\ref{tab:data_sources}.

\begin{table}[H]
  \caption{Composition of data sources in the research dataset.}
  \label{tab:data_sources}
  \centering
  \begin{tabular}{lcc}
    \toprule
    \textbf{Data Source} & \textbf{Number of Samples} & \textbf{Description} \\
    \midrule
    Student Opinions on AI & 1{,}154 & Main research data \\
    Positive and Negative Lexical Dictionary & 1{,}141 & Additional sentiment-based data \\
    \midrule
    \textbf{Total} & \textbf{2{,}295} & Final dataset \\
    \bottomrule
  \end{tabular}
\end{table}

The final dataset is nearly balanced, with 1{,}158 negative samples (50.46\%) and 1{,}137
positive samples (49.54\%).

The class distribution in the final dataset is presented in Table~\ref{tab:class_distribution}.

\begin{table}[ht]
  \caption{Class distribution in the final dataset.}
  \label{tab:class_distribution}
  \centering
  \begin{tabular}{lcc}
    \toprule
    \textbf{Class} & \textbf{Number of Samples} & \textbf{Percentage} \\
    \midrule
    Negative & 1{,}158 & 50.46\% \\
    Positive & 1{,}137 & 49.54\% \\
    \midrule
    \textbf{Total} & \textbf{2{,}295} & \textbf{100\%} \\
    \bottomrule
  \end{tabular}
\end{table}

\subsection{Data Cleaning and Preprocessing}
Preprocessing includes lowercasing, removal of URLs, mentions, hashtags, punctuation, and
extra whitespace, followed by normalization of non-standard words and repeated characters.

\subsection{Machine Learning Pipeline}
  
\begin{figure}[ht]
  \centering
  \includegraphics[width=0.68\textwidth]{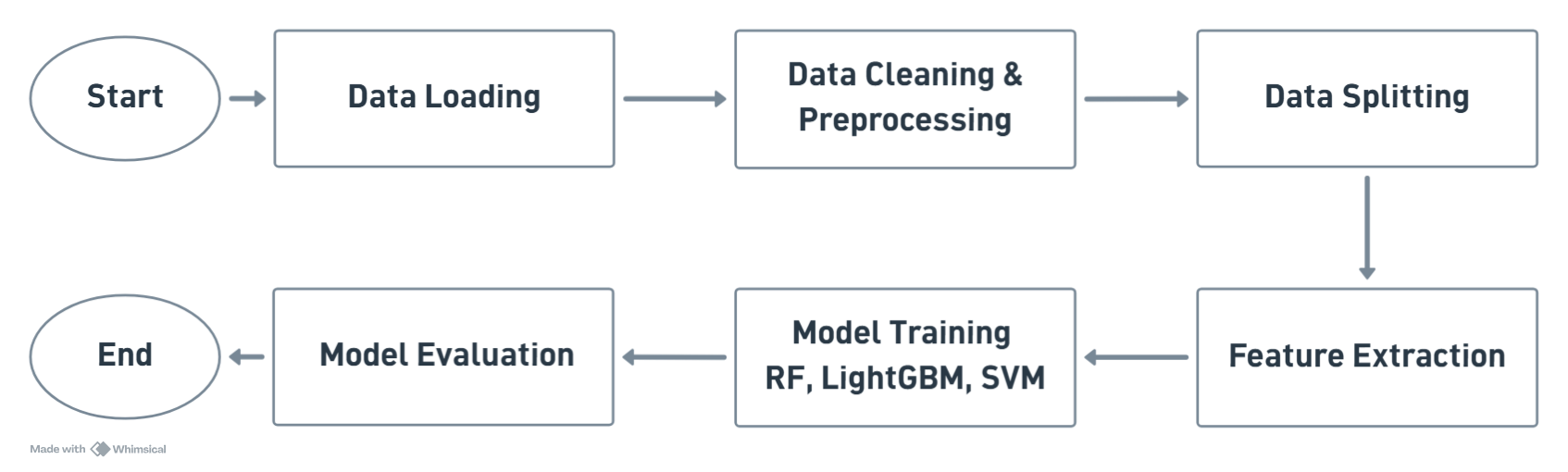}
  \caption{Machine learning pipeline for sentiment classification.}
  \label{fig:ml_pipeline}
\end{figure}

The machine learning pipeline includes preprocessing, splitting, TF-IDF feature extraction,
model training, selection, and evaluation, as shown in Figure~\ref{fig:ml_pipeline}.

After preprocessing, text is converted into TF-IDF features using unigrams and bigrams.
The vectorizer is fit only on training data to avoid leakage. LightGBM, Random Forest, and
SVM are compared on validation accuracy, with SVM selected for final testing and comparison
against DistilBERT.

\subsection{DistilBERT Architecture}
\label{sec:distilbert_architecture}

\begin{figure}[H]
  \centering
  \includegraphics[width=0.58\textwidth]{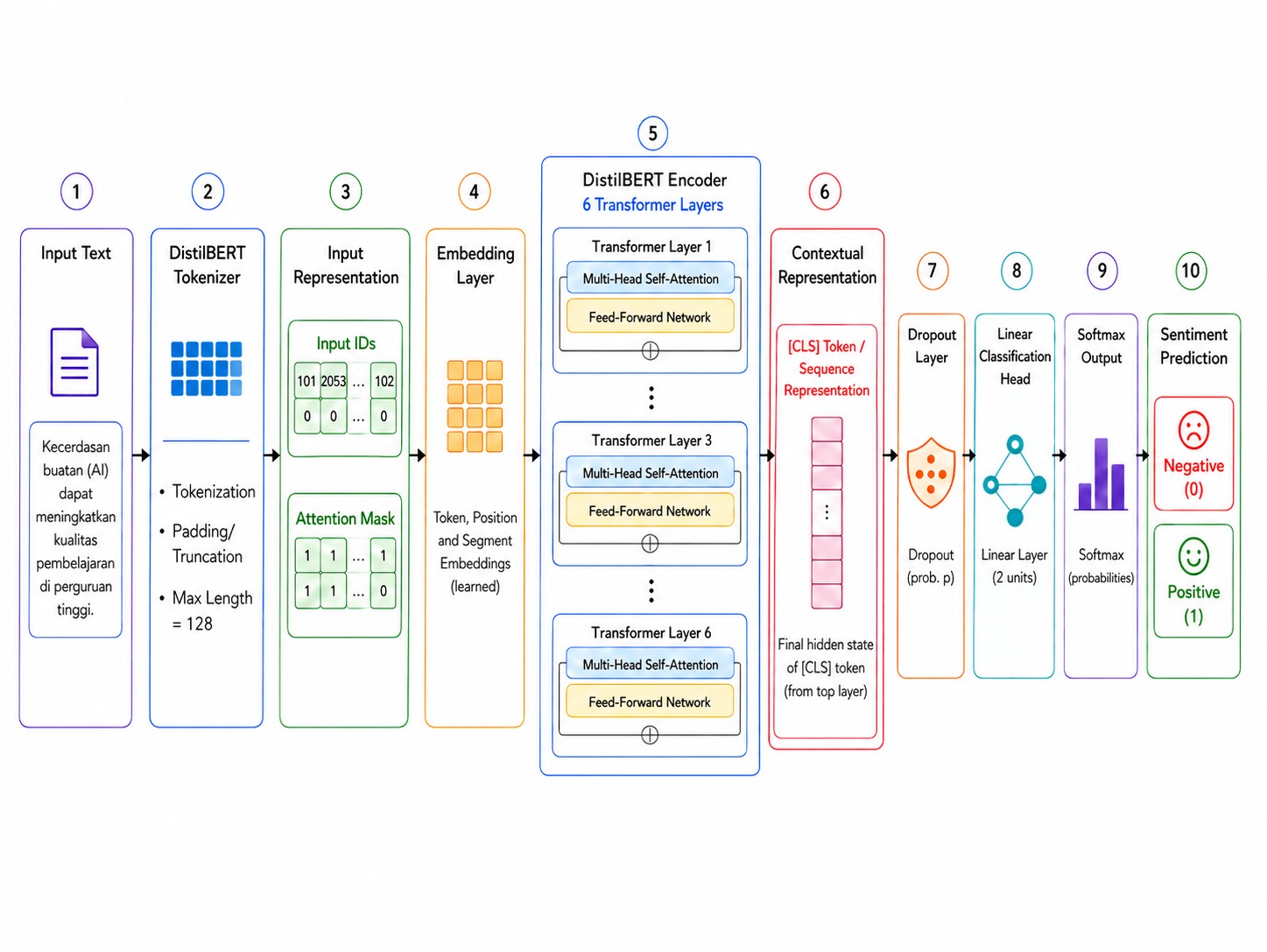}
  \caption{DistilBERT architecture for binary sentiment classification.}
  \label{fig:distilbert_architecture}
  \vspace{2pt}
  \footnotesize{Source: Visualization generated with the assistance of ChatGPT (OpenAI).}
\end{figure}

The deep learning approach uses DistilBERT, a lighter BERT variant, to capture contextual
information through self-attention. The architecture is shown in Figure~\ref{fig:distilbert_architecture}.

Input text is tokenized into input IDs and attention masks with a maximum length of 128,
processed by the six-layer DistilBERT encoder, and passed to a classification head for
negative or positive sentiment prediction.

\subsection{Experiment}
The experiment evaluates model performance on binary sentiment classification using the
preprocessed dataset.

\label{sec:experiment}

The cleaned dataset is used to classify texts into positive and negative classes. The dataset is split 80:10:10 into training, validation, and test sets, as shown in Table~\ref{tab:experiment_data_split}.

\begin{table}[H]
  \caption{Dataset split for the experiment.}
  \label{tab:experiment_data_split}
  \centering
  \begin{tabular}{lcc}
    \toprule
    \textbf{Subset} & \textbf{Number of Samples} & \textbf{Percentage} \\
    \midrule
    Training Data & 1{,}632 & 80\% \\
    Validation Data & 204 & 10\% \\
    Test Data & 459 & 10\% \\
    \bottomrule
  \end{tabular}
\end{table}

The machine learning comparison uses LightGBM, Random Forest, and SVM with the same TF-IDF features. Their configurations are listed in Table~\ref{tab:model_configuration}.

\begin{table}[H]
  \caption{Machine learning model configuration.}
  \label{tab:model_configuration}
  \centering
  \renewcommand{\arraystretch}{1.4}
  \resizebox{\textwidth}{!}{%
  \begin{tabular}{lllp{5.5cm}}
    \toprule
    \textbf{Model} & \textbf{Configuration} & \textbf{Main Parameter} & \textbf{Reason for Use} \\
    \midrule
    LightGBM &
    LGBMClassifier &
    Boosting ensemble &
    Fast and efficient as an initial baseline \\

    Random Forest &
    RandomForestClassifier &
    Parallel ensemble &
    Robust and able to reduce overfitting through averaging \\

    SVM &
    SVC(kernel='rbf', C=1.0) &
    Kernel-based classifier &
    Suitable for high-dimensional text data and non-linear separation \\
    \bottomrule
  \end{tabular}%
  }
\end{table}

LightGBM uses n\_estimators=100, max\_depth=10, and
learning\_rate=0.1; Random Forest uses n\_estimators=100,
max\_depth=20, and min\_samples\_split=5; and SVM uses an RBF kernel
with C=1.0. All models use random\_state=42 for reproducibility.

\begin{table}[H]
  \caption{Training configuration of the DistilBERT model.}
  \label{tab:distilbert_training_config}
  \centering
  \renewcommand{\arraystretch}{1.4}
  \begin{tabular}{lc}
    \toprule
    \textbf{Parameter} & \textbf{Value} \\
    \midrule
    Model & distilbert-base-uncased \\
    Max Length & 128 \\
    Vocabulary Size & 30.522 \\
    Batch Size & 64 \\
    Learning Rate & $1 \times 10^{-6}$ \\
    Epochs & 100 \\
    Early Stopping Patience & 5 \\
    Weight Decay & 0.001 \\
    Dropout & 0.5 \\
    Warmup Steps & 100 \\
    Optimizer & AdamW \\
    Scheduler & Linear \\
    Best Metric & Validation F1-score \\
    Device & CUDA \\
    \bottomrule
  \end{tabular}
\end{table}

Model performance is evaluated using accuracy, precision, recall, and F1-score, summarized in Table~\ref{tab:evaluation_metrics}.

\begin{table}[H]
  \caption{Evaluation metrics for the sentiment classification model.}
  \label{tab:evaluation_metrics}
  \centering
  \renewcommand{\arraystretch}{1.6}
  \resizebox{\textwidth}{!}{%
  \begin{tabular}{lll}
    \toprule
    \textbf{Metric} & \textbf{Formula} & \textbf{Description} \\
    \midrule
    Accuracy &
    $\frac{TP + TN}{TP + TN + FP + FN}$ &
    Proportion of correct predictions out of all data \\

    Precision &
    $\frac{TP}{TP + FP}$ &
    Model precision in predicting the positive class \\

    Recall &
    $\frac{TP}{TP + FN}$ &
    Model ability to recognize actual positive samples \\

    F1-score &
    $2 \times \frac{Precision \times Recall}{Precision + Recall}$ &
    Harmonic mean between precision and recall \\
    \bottomrule
  \end{tabular}%
  }
\end{table}

In Table~\ref{tab:evaluation_metrics}, $TP$, $TN$, $FP$, and $FN$ denote true positive,
true negative, false positive, and false negative predictions.

\section{Results and Discussion}
\label{sec:results}
\subsection{Dataset Summary and Data Split}

As a summary, the final dataset consists of 2{,}295 samples with a relatively balanced class distribution, namely 1{,}158 negative samples and 1{,}137 positive samples. The dataset is then divided into training data, validation data, and test data with a ratio of 80:10:10.


\label{sec:results_discussion}

\subsection{Machine Learning Model Results}

Three machine learning models LightGBM, Random Forest, and SVM are evaluated on the same TF-IDF features. Validation results are used for model selection, and test results for generalization assessment. \\

LightGBM. As a boosting-based baseline, LightGBM trains fastest but performs worst, with 57.35\% validation accuracy, 49.67\% test accuracy, and a 48.46\% F1-score. These results show that LightGBM is poorly suited to sparse, high-dimensional TF-IDF features. \\

Random Forest. Random Forest improves over LightGBM but remains below SVM, reaching 62.25\% validation accuracy and 58.61\% test accuracy with a 56.48\% F1-score.

Random Forest still handles TF-IDF features less effectively than SVM. \\

\begin{figure}[H]
  \centering
  \includegraphics[width=0.42\textwidth]{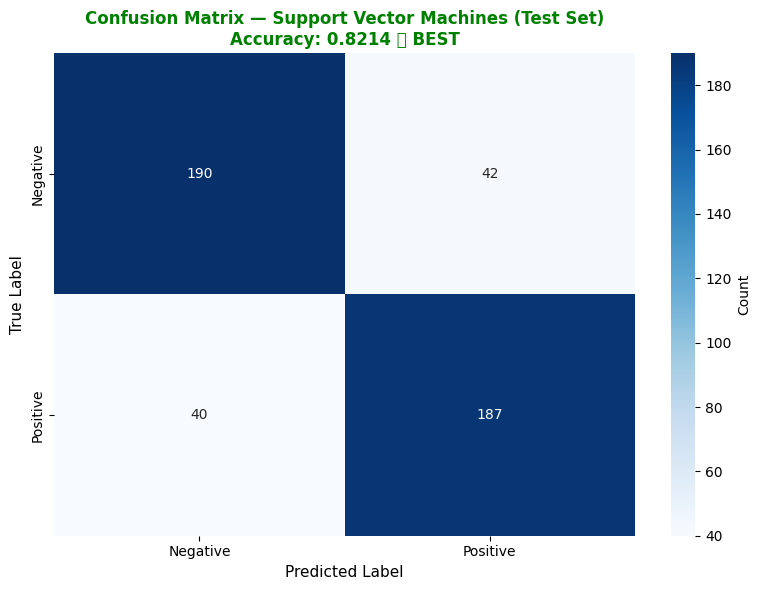}
  \caption{Confusion matrix of the SVM model on the test data.}
  \label{fig:cm_svm}
\end{figure}

SVM with an RBF kernel is the best machine learning model, achieving 87.25\% validation accuracy and 82.14\% test accuracy, precision, recall, and F1-score, as also reflected in Figure~\ref{fig:cm_svm}.

These results indicate that SVM provides the most balanced performance, which is consistent with its strength on high-dimensional TF-IDF features. \\

SVM correctly classifies 377 of 459 samples: 190 negatives and 187 positives, with 42 negative and 40 positive misclassifications. The error distribution is fairly balanced, indicating stable performance across both classes. \\

SVM achieves the best machine learning performance, with 87.25\% validation accuracy and 82.14\% test accuracy and F1-score. LightGBM performs worst, while Random Forest improves on LightGBM but remains below SVM, indicating that SVM is better suited to sparse,
high-dimensional TF-IDF features.

\subsection{DistilBERT Evaluation Results}

DistilBERT is used as the Transformer-based comparator because it models contextual
relationships more effectively than TF-IDF features. The experiment uses
\texttt{distilbert-base-uncased} with a maximum length of 128, batch size 64, learning rate
$1 \times 10^{-6}$, dropout 0.5, weight decay 0.001, AdamW, and early stopping with patience 5.

A per-class breakdown is given in Table~\ref{tab:distilbert_classification_report}. DistilBERT
attains an F1-score of 0.8548 on the negative class and 0.8402 on the positive class.

\begin{table}[H]
  \caption{Classification report of the DistilBERT model on the test data.}
  \label{tab:distilbert_classification_report}
  \centering
  \renewcommand{\arraystretch}{1.4}
  \begin{tabular}{lcccc}
    \toprule
    \textbf{Class} & \textbf{Precision} & \textbf{Recall} & \textbf{F1-score} & \textbf{Support} \\
    \midrule
    Negative & 0.8240 & 0.8879 & 0.8548 & 116 \\
    Positive & 0.8762 & 0.8070 & 0.8402 & 114 \\
    \midrule
    \multicolumn{3}{l}{Accuracy} & 0.8478 & 230 \\
    Macro Avg & 0.8501 & 0.8475 & 0.8475 & 230 \\
    Weighted Avg & 0.8499 & 0.8478 & 0.8475 & 230 \\
    \bottomrule
  \end{tabular}
\end{table}

Table~\ref{tab:distilbert_classification_report} shows balanced class performance. The model
is slightly more sensitive to negative samples and slightly more precise on positive predictions,
but the class F1-scores remain close.

\begin{figure}[!htbp]
  \centering
  \includegraphics[width=0.42\textwidth]{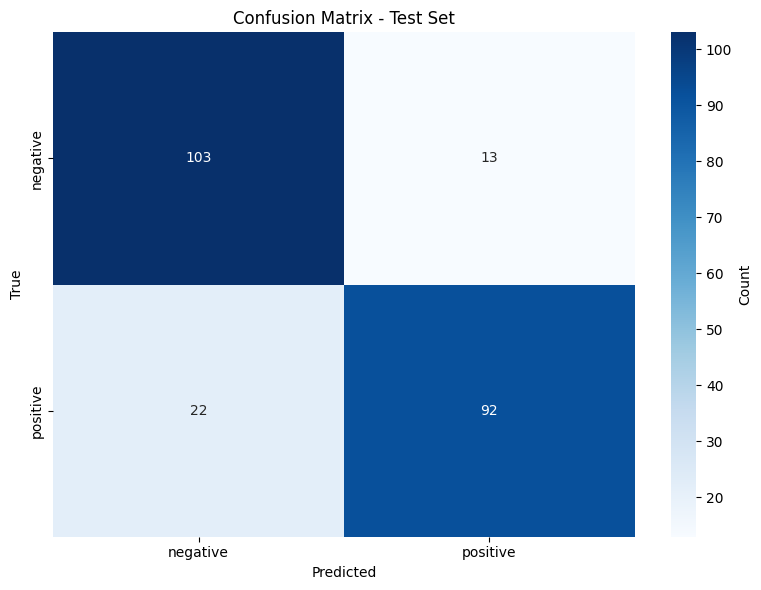}
  \caption{Confusion matrix of the DistilBERT model on the test data.}
  \label{fig:cm_distilbert}
\end{figure}

\begin{figure}[H]
  \centering

  \begin{subfigure}{0.40\textwidth}
    \centering
    \includegraphics[width=\textwidth]{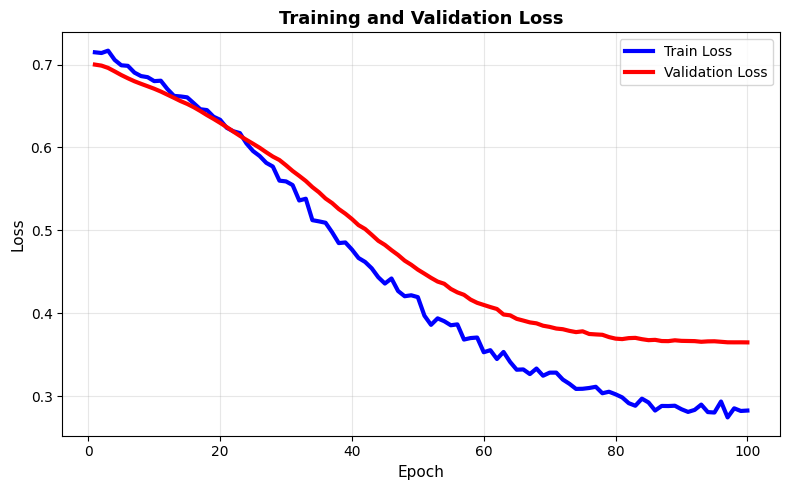}
    \caption{Loss}
    \label{fig:distilbert_loss}
  \end{subfigure}
  \hfill
  \begin{subfigure}{0.40\textwidth}
    \centering
    \includegraphics[width=\textwidth]{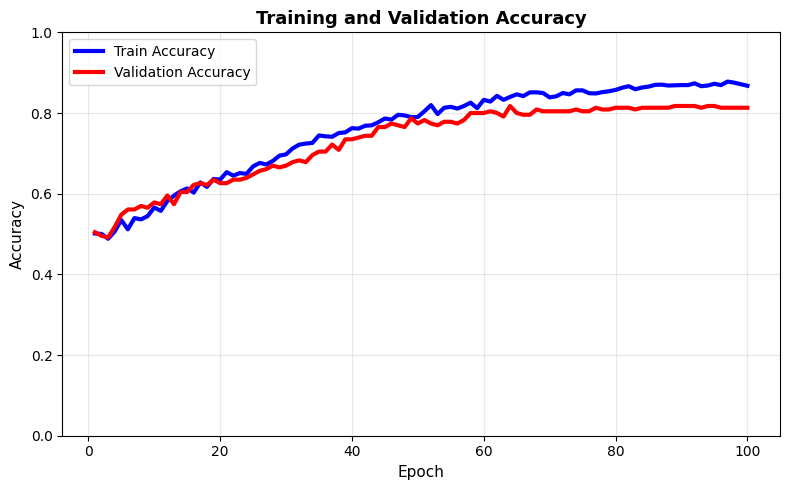}
    \caption{Accuracy}
    \label{fig:distilbert_accuracy}
  \end{subfigure}

\end{figure}

The final train and validation losses are 0.2827 and 0.3649, indicating only a modest gap.
The training curves and final test results (84.78\% accuracy and 84.75\% weighted F1-score)
confirm good generalization.

\subsection{Model Comparison}

Table~\ref{tab:model_comparison} compares the three machine learning models and DistilBERT.
Validation accuracy supports model selection, while test accuracy and F1-score indicate generalization.

\begin{table}[H]
  \caption{Performance comparison of machine learning and Transformer-based deep learning models on validation and test data.}
  \label{tab:model_comparison}
  \centering
  \renewcommand{\arraystretch}{1.4}
  \resizebox{\textwidth}{!}{%
  \begin{tabular}{lccccc}
    \toprule
    \textbf{Model} & \textbf{Approach} & \textbf{Val. Accuracy} & \textbf{Test Accuracy} & \textbf{Test F1-score} & \textbf{Training Time} \\
    \midrule
    LightGBM & Machine Learning & 57.35\% & 49.67\% & 48.46\% & 0.057s \\
    Random Forest & Machine Learning & 62.25\% & 58.61\% & 56.48\% & 0.216s \\
    SVM & Machine Learning & \textbf{87.25\%} & 82.14\% & 82.14\% & 0.210s \\
    DistilBERT & Transformer-based Deep Learning & 81.74\% & \textbf{84.78\%} & \textbf{84.75\%} & 22.8m \\
    \bottomrule
  \end{tabular}%
  }
\end{table}

SVM is the best machine learning model, while LightGBM and Random Forest lag behind on the
same TF-IDF features. DistilBERT delivers the best overall test performance, outperforming
SVM by 2.64\% in accuracy and 2.61\% in F1-score. However, SVM remains far more efficient to train.

\subsection{Deployment}

The best model is deployed as an interactive web application using Gradio and Hugging Face
Spaces, allowing users to submit opinion text and receive real-time sentiment predictions
with confidence scores.


\section{Conclusion}
\label{sec:conclusion}

This study analyzes Indonesian student opinions on AI use in higher education using TF-IDF-based machine learning and DistilBERT. Among the machine learning models, SVM performs best with 82.14\% test accuracy and F1-score, while DistilBERT achieves the best overall results with 84.78\% accuracy and 84.75\% F1-score. Thus, Transformer-based models better capture context, whereas SVM remains a strong low-cost alternative. Future work may expand the dataset, add a neutral class, use Indonesian-specific models such as IndoBERT, and explore aspect-based sentiment analysis.

\bibliographystyle{unsrtnat}
\bibliography{references}

\end{document}